\documentclass{article}
\usepackage[T1]{fontenc}
\usepackage{spconf,amsmath,graphicx}
\usepackage{booktabs,tabularx}
\usepackage[table]{xcolor}
\usepackage[hidelinks]{hyperref}
\usepackage{enumitem}
\usepackage{fancyhdr}

\newcommand{\headlogo}[2]{%
    \smash{\includegraphics[height=#1]{#2}}}
\fancypagestyle{paperfirst}{%
    \fancyhf{}%
    \fancyhead[L]{\headlogo{16pt}{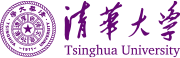}}%
    \fancyhead[R]{\headlogo{13pt}{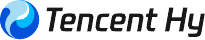}}%
    \fancyfoot[C]{\thepage}%
}
\fancypagestyle{paperrest}{%
    \fancyhf{}%
    \fancyfoot[C]{\thepage}%
}

\definecolor{SLBlue}{HTML}{0F4D92}
\definecolor{SLTint}{HTML}{EAF1F8}
\definecolor{SLGray}{HTML}{F2F3F5}

\newcommand{\trainable}{\textsuperscript{$\dagger$}}

\newcommand{\bbest}[1]{\textcolor{SLBlue}{\textbf{#1}}}
\newcommand{\obest}[1]{\underline{#1}}
\title{Thinking in Depth, Speaking Directly: Recurrent Latent Reasoning for
Paralinguistically Grounded Spoken Dialogue}

\name{\begin{tabular}{@{}c@{}}
    Shengbo Cai$^{1,3}$, Yuxiang Wang$^{2,3}$, Jingran Xie$^{1}$, Zhisheng Zhang$^{1}$, \\[2pt]
    Shun Lei$^{1}$, Di Cao$^{3}$, Teddy Sun$^{3}$, Zhiyong Wu$^{1,\dagger}$
    \end{tabular}
    \thanks{Work done during internship at Tencent Hunyuan.\\
    \hspace*{1.8em}$^{\dagger}$Corresponding author.}}
    \address{$^{1}$Tsinghua University \quad
    $^{2}$The Chinese University of Hong Kong, Shenzhen \quad
    $^{3}$Tencent Hunyuan}

\begin{document}
\ninept
\maketitle
\pagestyle{paperrest}
\thispagestyle{paperfirst}
\begin{abstract}
    Empathetic spoken dialogue requires models to use both what is said and how it is said to decide how to respond. Explicit CoT can improve paralinguistic perception and make acoustic cues more explicit in replies, yet does not ensure their effective use in response planning.
    We call this mismatch the \emph{perception--reasoning gap}. In addition, CoT may not fully capture acoustic cues in words, and generating it adds inference latency. To address these limitations, we introduce \textbf{LoopSLM}, which builds on looped Transformers for latent reasoning, reusing a decoder block to refine hidden states with acoustic grounding at every pass. Its two-stage training further narrows the \emph{perception--reasoning gap} by separating learning to reason from learning to respond, enabling direct inference without CoT. On EchoMind, LoopSLM improves
    paralinguistic understanding,
    reasoning, and reply quality over Qwen2.5-Omni-7B. Against the CoT-SFT baseline, LoopSLM gains over 20 points in reasoning accuracy while generating 64.5\% fewer tokens at half the latency. It also outperforms Qwen3-Omni-Thinking on most
    empathetic reply metrics with $34\times$ lower latency. Despite training only
    on dialogue data, LoopSLM improves accuracy on general audio benchmarks.
    \end{abstract}
\begin{keywords}
Speech language models, paralinguistic reasoning, empathetic dialogue, looped Transformers, latent reasoning
\end{keywords}
\section{Introduction}
\label{sec:intro}
In empathetic spoken dialogue, what a speaker conveys and what constitutes
an appropriate reply depend not only on the words but also on how
they are delivered. The same words can convey confidence, hesitation,
frustration, or vulnerability through prosody, voice quality, and timing.
A speech language model (SLM) must therefore do more than recognize these
paralinguistic cues. It must infer what they imply about the speaker's state
and communicative needs, then use that inference to shape its
reply~\cite{Zhou26EchoMind,Wang26ParaBridge}. Paralinguistically grounded
reasoning links \emph{how} something is said to \emph{what} the model should
say in response. The challenge is to provide supervision that teaches models how to use paralinguistic cues when planning a reply.

Chain-of-thought (CoT) provides one such supervisory signal by making
intermediate reasoning explicit~\cite{Wei22CoT}. Speech-grounded CoT can encourage models to attend to acoustic evidence~\cite{Tian25StepAudio}.
Yet recognizing and describing that evidence does not ensure that models can
use it for reasoning. In our comparisons, CoT fine-tuning improves
paralinguistic understanding and makes generated replies more explicit about
these cues but weakens models' ability to use them when deciding how to
respond (Table~\ref{tab:main_results}). We call this mismatch the
\emph{perception--reasoning gap}. Beyond this reasoning gap, textual CoT also
increases inference latency, as reasoning tokens must be generated before
the reply. Can CoT be used only during training to teach the model to reason with paralinguistic cues while allowing direct responses at inference?

Latent reasoning avoids textual CoT at inference but does not by itself close
the \emph{perception--reasoning gap}. Models therefore still need
supervision on how paralinguistic cues should shape replies. For example,
Coconut and CODI learn latent steps through gradual CoT replacement or
representation distillation, leaving those steps only indirectly aligned
with the original CoT and weakening credit assignment across the reasoning
process~\cite{Hao25Coconut,Shen25CODI}. CoLaR addresses this limitation with
denser CoT targets, and its gains confirm that supervision quality is
critical~\cite{Tan25CoLaR}. Yet it still predicts compressed latent steps
autoregressively at inference. Together, these findings motivate a design
that combines dense speech-grounded CoT supervision with iterative reasoning
through recurrent decoder depth. Looped Transformers, in turn, supply the
recurrent-depth component by reusing layers for iterative refinement of
hidden states, an approach shown to improve reasoning without adding latent
positions~\cite{Saunshi25Looped,Geiping25Recurrent,Wang26RecurTrace}.

We therefore introduce \textbf{LoopSLM}, which combines recurrent reasoning
in decoder depth with acoustic grounding at every pass. Two-stage training
first uses speech-grounded CoT to strengthen paralinguistic understanding
and reasoning, then uses the refined states to improve empathetic replies
while preserving that reasoning. LoopSLM responds directly at inference
without CoT.

Relative to Qwen2.5-Omni-7B, LoopSLM improves mean paralinguistic
understanding and reasoning accuracy on EchoMind by 9.8 and 6.6 points,
respectively, and raises scores in all four dimensions of reply quality.
Against a matched CoT-SFT baseline, it gains more than 20 points in reasoning
accuracy with 64.5\% fewer generated tokens and half the latency. It also
outperforms Qwen3-Omni-Thinking~\cite{Xu25Qwen3Omni} on most metrics of
empathetic reply quality with $34\times$ lower latency. Although trained only
on dialogue data, it improves accuracy on the general audio benchmarks MMSU
and MMAU-Pro. Our contributions are:

\begin{figure*}[!ht]
    \centering
    \includegraphics[width=\textwidth]{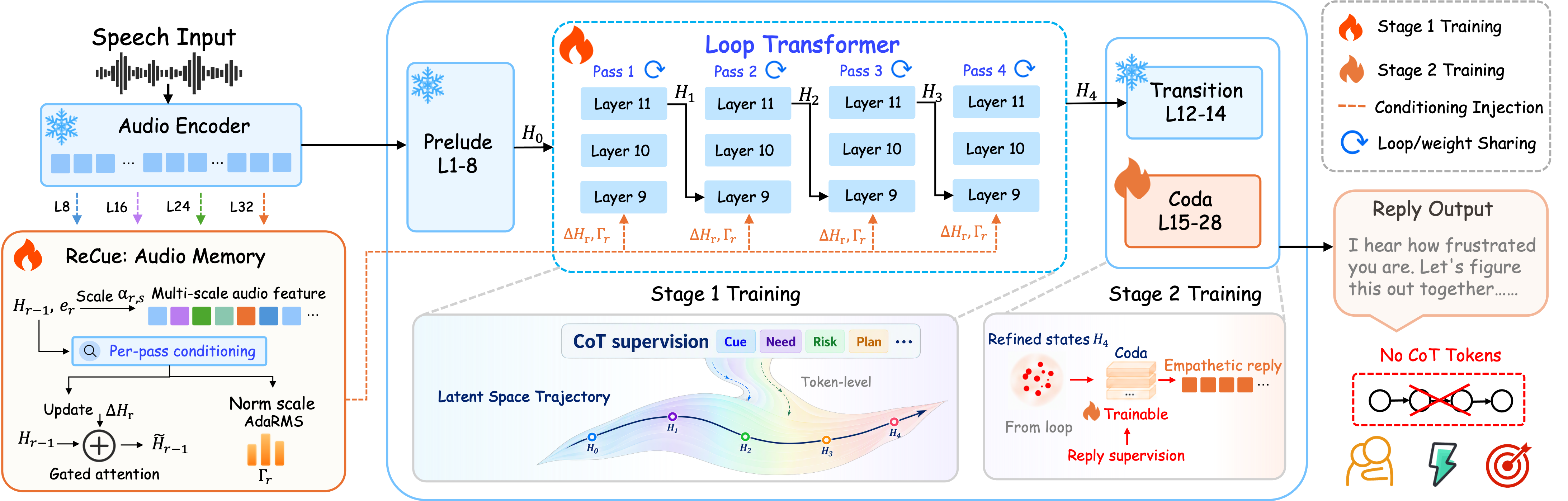}
    \caption{LoopSLM overview. Speech-grounded CoT trains a weight-shared Loop
unrolled for four passes, with ReCue re-anchoring each pass to acoustic
memory. The Loop and ReCue are then frozen while the Coda learns a
direct-response readout. Inference emits no CoT tokens.}
    \label{fig:overview}
\end{figure*}

\begin{itemize}[leftmargin=1.2em, itemsep=0pt, parsep=0pt, topsep=0pt]

\item We propose \textbf{LoopSLM}, which, to our knowledge, is the first SLM
to incorporate a looped Transformer architecture. It moves latent response planning to recurrent decoder depth without autoregressive reasoning tokens at inference.

\item We introduce a \emph{two-stage CoT-to-response curriculum} that
separates learning to reason from learning to respond. Stage~1 strengthens
paralinguistic reasoning with speech-grounded CoT, while Stage~2 turns the
refined states into empathetic replies.

\item We provide controlled evidence that LoopSLM's gains depend on learned recurrent refinement rather than depth alone, while isolating the roles of acoustic memory and staged supervision.

\end{itemize}

\section{Method}
\label{sec:method}

LoopSLM moves speech-grounded response planning from the sequence axis to
recurrent decoder depth without adding latent positions or reasoning tokens
(Fig.~\ref{fig:overview}). The architecture combines a shared early-middle
Loop for recurrent refinement, \textbf{ReCue} (\textbf{Re}-anchoring to acoustic
\textbf{Cue}s) for acoustic grounding at every pass,
and a Coda for direct response generation. Let $H_0$ denote the state
entering the Loop and $H_r$ the state after pass $r$, for
$r=1,\ldots,R$. Training separates recurrent refinement under CoT
supervision from direct response learning.

\subsection{ReCue: per-pass acoustic re-anchoring}
\label{sec:bridge}

Prior layerwise analyses show that paralinguistic information is strongly
encoded by audio encoders but can degrade across the decoder, including in
Qwen2.5-Omni~\cite{Koduru26Heard}. Our analysis further finds that decoder
representations of identical words with different delivery lose much of
their separation before the middle layers. Placing the Loop early lets
refinement begin while more of this information remains, but each recurrent
pass still receives only the previous decoder state. ReCue therefore gives
every pass renewed access to the original acoustic evidence through a
compact memory built from multiple depths of the frozen audio encoder.

Let $A_s$ denote the encoder activations at depth $s$, $\Psi_s$ their
projection, and $Q$ a shared set of 16 learned queries. We write
$\mathrm{CA}(q,m)$ for cross-attention from $q$ to $m$. One bank per scale,
\begin{equation}
M_s=\mathrm{CA}\big(Q,\Psi_s(A_s)\big),\quad s\in\{8,16,24,32\},
\label{eq:memory}
\end{equation}
forms the memory $M=\{M_s\}$, built once per utterance. Pass $r$ then mixes
the scales and reads the result into the state:
\begin{align}
\widehat M_r&=\textstyle\sum_s\alpha_{r,s}M_s,\nonumber\\
\widetilde H_{r-1}&=H_{r-1}+g_r\,\mathrm{CA}\big(H_{r-1}+e_r,\widehat M_r\big),
\label{eq:inject}
\end{align}
with learned mixing weights $\alpha_r$ conditioned on the state pooled over
audio positions and on an iteration embedding $e_r$ that distinguishes
passes. A second head reads the same inputs and produces the scale-only
modulation $\Gamma_r$ of the Loop's pre-norms; ReCue therefore returns
$(\widetilde H_{r-1},\Gamma_r)=C_\phi(H_{r-1},M,r)$ with parameters $\phi$.
The gate $g_r$ and the $\Gamma_r$ head are zero-initialized, so ReCue
leaves the backbone unperturbed at initialization.

\begin{table*}[!ht]
    \centering
    \caption{Paralinguistic understanding, reasoning, empathetic replies, and
    general audio. S/H: synthetic/human speech; D/G: DeepSeek/Gemini judges.
    \bbest{Blue bold}: best among comparable-scale ($\le$9B) systems;
    \obest{underline}: best overall.}
    \label{tab:main_results}
    \vspace{3pt}
    \begingroup
\fontsize{9}{10.5}\selectfont
\setlength{\tabcolsep}{2.2pt}
\renewcommand{\arraystretch}{1.08}
\begin{tabularx}{\textwidth}{>{\raggedright\arraybackslash}Xcr*{6}{c}*{3}{r}}
\toprule
& \multicolumn{2}{c}{\textbf{Cost}} & \multicolumn{7}{c}{\textbf{Paralinguistic dialogue}} & \multicolumn{2}{c}{\textbf{General audio}} \\
\cmidrule(lr){2-3}\cmidrule(lr){4-10}\cmidrule(l){11-12}
& & & \multicolumn{2}{c}{Accuracy (S/H)} & \multicolumn{4}{c}{Response (D/G)} & & & \\
\cmidrule(lr){4-5}\cmidrule(lr){6-9}
Model & \#P & Tok$\downarrow$ & Underst. & Reason. & C1 & C2 & C3 & C4 & SD-Eval & MMSU & MMAU-Pro \\
\midrule
\rowcolor{SLGray}
\multicolumn{12}{l}{\itshape Off-the-shelf models} \\
Qwen2.5-Omni & 7B & 43.5 & 60.47/55.09 & 57.59/56.60 & 4.57/4.42 & 4.31/4.41 & 4.00/4.40 & 1.37/1.46 & 4.62 & 63.46 & 53.71 \\
Kimi-Audio & 7B & 25.3 & 46.31/43.58 & 50.78/49.31 & 3.45/3.14 & 2.82/2.73 & 2.38/2.63 & 1.72/1.72 & 2.86 & 59.98 & 45.91 \\
Audio-Flamingo-3 & 7B & 18.2 & 65.08/55.30 & 59.47/58.17 & 1.96/1.40 & 1.37/1.22 & 1.92/1.84 & 1.23/1.09 & 2.30 & 61.64 & 53.20 \\
OSUM-EChat & 3B & 120.5 & 40.37/37.07 & 50.33/52.32 & 3.80/3.31 & 3.73/3.60 & 4.16/4.12 & 1.84/1.96 & 3.34 & 51.60 & 42.97 \\
MiniCPM-o 4.5 & 9B & 17.5 & 67.19/58.96 & 63.25/62.57 & 2.70/2.21 & 2.11/2.00 & 2.92/2.87 & 1.30/1.24 & 1.77 & 58.36 & 50.29 \\
MiMo-Audio-Think\textsuperscript{$\dagger$} & 7B & 377.8 & 56.02/47.05 & 60.88/60.24 & 3.89/3.23 & 2.78/2.17 & 2.63/2.31 & 2.01/1.94 & 4.88 & 61.92 & 54.45 \\
\addlinespace[2pt]
Qwen3-Omni-Thinking\textsuperscript{$\dagger$} & 30B & 711.0 & 69.13/70.06 & 61.85/66.52 & 4.64/4.35 & 4.02/3.72 & 3.96/3.84 & \obest{2.76}/2.79 & 4.82 & 68.98 & \obest{61.82} \\
\rowcolor{SLGray}
\multicolumn{12}{l}{\itshape CoT-SFT on our data (CoT emission not enforced)} \\
Qwen2.5-Omni & 7B & 96.9 & 60.86/56.62 & 43.77/42.46 & 4.78/4.70 & 4.52/4.71 & 4.05/4.46 & \bbest{2.60}/2.81 & 4.83 & 63.74 & 52.77 \\
Kimi-Audio & 7B & 81.5 & 56.73/50.31 & 42.50/37.12 & 4.68/4.69 & 4.31/4.66 & 3.79/4.37 & \bbest{2.60}/\bbest{\obest{2.83}} & 5.10 & 56.98 & 46.05 \\
Audio-Flamingo-3 & 7B & 32.3 & 70.05/57.03 & 54.87/54.27 & 4.55/4.53 & 4.15/4.52 & 3.61/4.22 & 1.68/1.94 & 5.14 & 59.96 & 53.09 \\
\addlinespace[2pt]
Qwen3-Omni-Thinking\textsuperscript{$\dagger$} & 30B & 171.6 & \obest{72.78}/\obest{70.77} & \obest{69.55}/\obest{70.92} & 4.75/4.71 & 4.40/4.64 & 3.80/4.33 & 2.59/2.79 & 4.87 & \obest{69.82} & 60.05 \\
\rowcolor{SLGray}
\multicolumn{12}{l}{\itshape LoopSLM on Qwen2.5-Omni-7B (direct response)} \\
LoopSLM-R1 & 7B & 35.2 & 67.33/61.91 & 60.80/58.67 & 4.74/4.68 & 4.59/4.73 & 4.10/\bbest{\obest{4.59}} & 1.93/2.02 & 5.19 & 63.72 & 49.29 \\
\rowcolor{SLTint}
\textcolor{SLBlue}{\textbf{LoopSLM-R4}} & 7B & 34.4 & \bbest{70.10}/\bbest{64.97} & \bbest{63.90}/\bbest{63.38} & \bbest{\obest{4.80}}/\bbest{\obest{4.75}} & \bbest{\obest{4.62}}/\bbest{\obest{4.76}} & \bbest{\obest{4.21}}/4.54 & 2.24/2.34 & \bbest{\obest{5.22}} & \bbest{64.72} & \bbest{56.09} \\
\quad w/o ReCue mem. & 7B & 34.8 & 65.52/61.41 & 63.34/62.63 & 4.67/4.62 & 4.39/4.66 & 3.93/4.45 & 2.26/2.44 & 4.73 & 63.44 & 53.98 \\
\bottomrule
\end{tabularx}
\endgroup

    \par\smallskip
    \begin{minipage}{\textwidth}
    \fontsize{9}{10.5}\selectfont
    \#P denotes total parameters. Tok is the mean continuation length in each
model's native tokens and includes emitted reasoning. $\dagger$ denotes
native thinking. Qwen3-Omni is an A3B-30B MoE.
    \end{minipage}
    \vspace{-2pt}
\end{table*}

\subsection{Weight-shared looped Transformer}
\label{sec:recurrent_depth}

To increase computation per token without extending the sequence, we
repeat a compact block of decoder
layers~\cite{Saunshi25Looped,Geiping25Recurrent}. We build on the 28-layer
Qwen2.5-Omni-7B Thinker~\cite{Xu25QwenOmni} and select the recurrent block
and response readout boundary through placement experiments. Looping
L9--11 rather than L15--17 improves understanding and reasoning on EchoMind
by 6.29 and 3.57 points, respectively, whereas L15--17
improves response style. This late-layer pattern echoes textual CoT by
favoring response expression over reasoning. At the other extreme, looping
L3--5 harms
transcription fidelity.

Together, these results suggest a division of labor along decoder depth.
L9--11 occupies an early-middle region where transcription remains stable
and vocal distinctions remain accessible, whereas layers from L15 onward
mainly support response expression. Accordingly, we partition the decoder
into a Prelude (L1--8), the Loop (L9--11, $\theta_L$), a Transition
(L12--14), and the Coda (L15--28, $\theta_C$). The Prelude and Transition
remain frozen throughout training.

The Loop input $H_0$ is the output of L8. At pass $r$, the shared block
updates the state returned by ReCue as
\begin{equation}
H_r = F_{9:11}(\widetilde H_{r-1};\theta_L,\Gamma_r),
\quad r=1,\ldots,R.
\label{eq:recurrence}
\end{equation}
All passes share $\theta_L$ and operate on the previous pass's state. Each
pass maintains a separate causal attention history over earlier tokens, so
recurrence increases depth without breaking autoregressive causality. The
Transition and Coda map $H_R$ to the next-token distribution through the
original language model head.
We use $R=4$ throughout training. At inference, the four passes run during
prompt processing and at each decoding step, increasing executed depth from
28 to 37 layer calls without duplicating Transformer weights.

\subsection{Two-stage CoT-to-response training}
\label{sec:two_stage_training}

CoT and response supervision serve different roles, so LoopSLM applies them
in separate stages. For user speech $a$ and task context $x$, each training
example provides a speech-grounded CoT trace $z$ and a response $y$
(Sec.~\ref{sec:training_data}). Both stages minimize the same token-level
cross-entropy over a supervised span $u_k$.
\begin{equation}
\mathcal L_k=-\frac{1}{|u_k|}\sum_{t=1}^{|u_k|}
\log p_\Theta(u_{k,t}\mid a,x,u_{k,<t}).
\label{eq:stage_loss}
\end{equation}
Here, $\Theta$ contains all model parameters, while
$\Omega_k\subset\Theta$ is the subset updated in stage $k$.

\noindent\textbf{Stage 1. CoT-supervised recurrent refinement.}
Recurrence adds computation but does not by itself teach how acoustic
evidence should shape the reply. Stage 1 sets $u_1=z$ and
$\Omega_1=\{\theta_L,\phi\}$, so only the Loop and ReCue are updated.
With the Coda fixed, this loss directs CoT supervision into the recurrent
path and teaches it to transform acoustic evidence into a response plan.

\noindent\textbf{Stage 2. Direct-response readout.}
Stage 2 starts from the Stage 1 model and sets $u_2=y$ and
$\Omega_2=\{\theta_C\}$ in~\eqref{eq:stage_loss}, where $\theta_C$ includes
the final normalization. Using response targets with no preceding trace, we
train only the Coda to map the refined states directly to replies and leave
the Loop and ReCue frozen. This preserves the recurrent refinement learned
from CoT.

\section{Experimental Setup}
\label{sec:experimental_setup}

\subsection{Training data}
\label{sec:training_data}

Stage~1 bootstraps the recurrent path from explicit CoT, so each trace must
explain how vocal cues affect the reply rather than restating the
transcript. Since existing corpora lack such supervision, we construct a
bilingual dataset from re-annotated LIME-440K~\cite{Zhao26Beyond} and public
human-recorded emotional-speech corpora. Gemini-3.5-Flash annotates each clip
with a compact \emph{Cue / Need / Risk / Plan} trace and a response:
\emph{Cue} captures audible prosodic and voice-quality evidence, while the
remaining slots encode the user's need, response risks, and reply strategy.
We retain only answerable conversational utterances whose affect is not
explicit in the transcript, reducing text-only shortcuts. Because Stage~1
distills explicit CoT into latent reasoning, we keep the traces at roughly 70 tokens. After filtering, the dataset contains 380,540 utterances (431.9\,h), with 71\% synthetic audio by duration and a 62.2/37.8\% Chinese/English split. We train on eight H800 GPUs, running Stage~1 for one epoch over the full dataset and Stage~2 for half an epoch.

\subsection{Evaluation}
\label{sec:evaluation}

EchoMind~\cite{Zhou26EchoMind} evaluates paralinguistic understanding,
reasoning, and empathetic reply generation. We report understanding/reasoning
MCQ accuracy (\%) on synthetic/human speech (S/H). DeepSeek V4 Flash and
Gemini-3.7-Flash (D/G) rate replies on a 1--5 scale for context fit (C1),
naturalness (C2), colloquialism (C3), and speech information relevance (C4).
SD-Eval~\cite{Ao24SDEval} provides DeepSeek scores on a 1--10 scale for
dialogues spanning emotion, accent, age, and background sound.
MMSU~\cite{Wang26MMSU} and MMAU-Pro~\cite{Kumar26MMAUPro} assess general
audio understanding and reasoning using accuracy (\%).

We compare off-the-shelf models, CoT-SFT models trained on our data, and
LoopSLM. The five additional off-the-shelf baselines are Kimi-Audio,
Audio-Flamingo-3, OSUM-EChat, MiniCPM-o 4.5, and
MiMo-Audio~\cite{KimiTeam25KimiAudio,Goel25AudioFlamingo3,
Geng25OSUMEChat,Cui26MiniCPMO,Zhang25MiMoAudio}.
The Qwen2.5 comparison uses the same base model and training data for CoT-SFT
and LoopSLM. CoT emission is not enforced for CoT-SFT at inference.
Cross-model comparisons use median end-to-end latency for naturally terminated
continuations. Figure~\ref{fig:compute_ablation}(a) reports median fixed-length
latency for generating 64 continuation tokens at batch size 1 on a single H800 GPU.

\section{Results}
\label{sec:results}

\subsection{Stronger reasoning with empathetic replies}

LoopSLM improves paralinguistic understanding, reasoning, and empathetic
reply quality (Table~\ref{tab:main_results}). Relative to
Qwen2.5-Omni-7B, LoopSLM-R4 raises mean understanding and reasoning accuracy
by 9.8 and 6.6 percentage points, respectively, and improves every reply
quality metric. Among systems with $\leq9$B parameters, LoopSLM achieves the
best performance on most evaluation metrics. It also surpasses
Qwen3-Omni-Thinking on understanding and reasoning for synthetic speech and
on most reply quality metrics, with $34\times$ lower median latency.
Although trained only on dialogue data, LoopSLM improves both general audio
benchmarks, showing transfer beyond its training domain.

Across all three non-thinking backbones, CoT-SFT improves paralinguistic
understanding and speech information relevance (C4) but lowers reasoning
accuracy on synthetic and human speech, revealing the
\emph{perception--reasoning gap}. By contrast, the native-thinking Qwen3-Omni benefits from structured CoT supervision, which improves
reasoning and shortens continuations, indicating that it can refine an
established thinking mode. LoopSLM retains the same supervision without
generating CoT at inference. Relative to the matched Qwen2.5 CoT-SFT
baseline, it gains over 20 percentage points in reasoning accuracy with
64.5\% fewer generated tokens. Although CoT-SFT retains a higher C4 score,
LoopSLM improves understanding, C4, and reasoning over the base model. It therefore
narrows this gap by moving CoT-supervised computation from the output
sequence to recurrent decoder depth.

\begin{table}[t]
\centering

\caption{Training ablations on EchoMind.
Bold indicates the best result ($R{=}4$). S$k$ denotes stage $k$, and \trainable{} indicates that the Loop remains trainable in S2. S/H and C1--C4 follow Sec.~\ref{sec:evaluation}.}
\label{tab:two_stage}
\vspace{3pt}
\begingroup
\fontsize{9}{10.0}\selectfont
\setlength{\tabcolsep}{1.8pt}
\renewcommand{\arraystretch}{1.00}
\begin{tabular}{@{}l>{\hspace{1.2pt}}r>{\hspace{0.8pt}}r>{\hspace{1.6pt}}rrrr>{\hspace{1.6pt}}r@{}}
\toprule
& \multicolumn{2}{c}{Accuracy (S/H)} & \multicolumn{4}{c}{Response (D)} & \\
\cmidrule(lr){2-3}\cmidrule(lr){4-7}
Training recipe & Under. & Reason. & C1 & C2 & C3 & C4 & Tok$\downarrow$ \\
\midrule
\rowcolor{SLGray}
\multicolumn{8}{@{}l}{\itshape Two-stage training ablations, all at $R{=}4$} \\
w/o S2 & 63.0/59.1 & 61.1/59.3 & 2.95 & 2.42 & 2.47 & 1.70 & 195.0 \\
Joint CoT+Resp. & 66.4/59.7 & 51.5/49.6 & 3.57 & 3.31 & 3.06 & 1.70 & 70.7 \\
S1, S2 Loop\,\trainable & 69.6/63.7 & \textbf{65.1}/\textbf{63.5} & 4.69 & 4.44 & 4.04 & 2.26 & 35.1 \\
Resp.$\to$Resp. & 65.4/60.2 & 56.8/54.3 & 4.71 & 4.36 & 3.84 & \textbf{2.30} & \textbf{34.0} \\
\addlinespace[1pt]
\rowcolor{SLTint}
\textcolor{SLBlue}{\textbf{LoopSLM-R4}} & \textbf{70.1}/\textbf{65.0} & 63.9/63.4 & \textbf{4.80} & \textbf{4.62} & \textbf{4.21} & 2.24 & 34.4 \\
\addlinespace[2pt]
\rowcolor{SLGray}
\multicolumn{8}{@{}l}{\itshape Trained loop depth} \\
$R{=}1$ & 67.3/61.9 & 60.8/58.7 & 4.74 & 4.59 & 4.10 & 1.93 & 35.2 \\
$R{=}2$ & 70.2/63.5 & 62.7/61.2 & 4.78 & 4.60 & 4.16 & 1.94 & 35.5 \\
$R{=}3$ & 69.0/62.9 & 63.3/60.9 & 4.79 & 4.56 & 4.05 & 2.05 & 34.8 \\
$R{=}6$ & 47.5/23.4 & 57.5/49.8 & 4.74 & 4.41 & 4.08 & 2.51 & 36.8 \\
\bottomrule
\end{tabular}
\endgroup

\end{table}

\subsection{What turns CoT supervision into latent reasoning?}

Table~\ref{tab:two_stage} shows that transferring CoT supervision into
latent reasoning requires a CoT objective in Stage~1 and separate response
learning in Stage~2. Replacing the Stage~1 CoT targets with response targets
keeps outputs short but lowers reasoning accuracy, confirming that gains
come from reasoning supervision rather than learning a concise response
format. Stage~1 alone retains much of the reasoning benefit but produces
long, poorly rated replies, whereas Stage~2 reduces output length by a
factor of 5.7 and improves reasoning. Joint training on CoT and
responses underperforms, confirming that recurrent refinement and response
readout should be learned separately.

Leaving the Loop trainable in Stage~2 yields little additional reasoning
accuracy but lowers understanding and most reply scores, showing that
response learning should preserve the refinement learned from CoT. Removing
ReCue memory mainly lowers understanding and reply quality
(Table~\ref{tab:main_results}), confirming the value of acoustic grounding.
These ablations show that LoopSLM turns CoT supervision into recurrent
reasoning grounded in acoustic evidence, enabling high-quality direct
responses without CoT at inference.

\begin{figure}[t]
\centering
\includegraphics[width=\columnwidth]{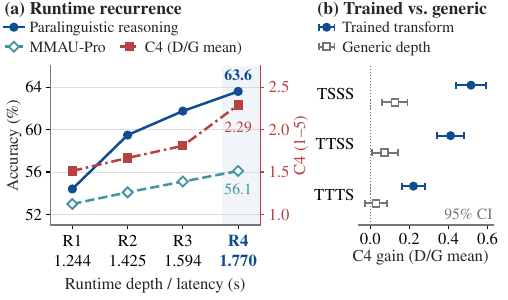}
\caption{Recurrent computation in LoopSLM-R4.
(a) R$k$ runs $k$ of the four trained passes. (b) Rows keep the leading passes trained (T) and fill the rest
with stock (original L9--11) blocks (S), with ReCue active. Generic depth $=$
C4(row) $-$ C4(S passes skipped); trained transform $=$ C4(TTTT) $-$
C4(row), on synthetic speech. C4: D/G mean; error bars: 95\%
item-paired bootstrap CIs.}

\label{fig:compute_ablation}
\end{figure}

\subsection{Is more depth enough for latent reasoning?}
\label{sec:compute_results}

More recurrent computation helps, but depth alone does not explain
LoopSLM's gains. Within LoopSLM-R4, executing more trained passes improves
reasoning and the use of speech information at higher inference cost
(Fig.~\ref{fig:compute_ablation}(a)). Because this sweep truncates a model
trained for four passes, it mixes the effect of additional computation with
a mismatch between training and inference depth. We therefore use the block
replacement control at fixed depth in Fig.~\ref{fig:compute_ablation}(b) to
isolate the effect of generic depth.

Stock blocks yield little improvement over skipping those passes, whereas
replacing more of them with trained recurrent blocks at the same depth
produces progressively larger C4 gains. These results show that the benefit comes from the learned
recurrent transformation rather than additional layer evaluations alone.
Models trained at different recurrence depths show that the benefit is not monotonic (Table~\ref{tab:two_stage}, bottom). R4 gives the strongest overall reasoning. Although R6 raises C4, degraded format adherence contributes to its lower reasoning accuracy and most reply quality scores. Effective latent reasoning therefore depends on both a learned recurrent transformation and an appropriate depth that maintains format adherence.

\section{Conclusion}
\label{sec:conclusion}
To our knowledge, LoopSLM is the first SLM to use looped Transformer
recurrence. It transfers speech-grounded CoT supervision into acoustically
grounded recurrent refinement in decoder depth. Empirically, it improves
paralinguistic reasoning and reply quality while avoiding autoregressive CoT
generation, and ablations show that these gains come from learned refinement
rather than depth alone. By improving both paralinguistic understanding and
reasoning relative to its backbone, LoopSLM narrows the
\emph{perception--reasoning gap} exposed by conventional CoT fine-tuning.

\section{Compliance with Ethical Standards}
This work collected no new data from human participants or animals. Training
used an internally curated corpus derived exclusively from appropriately
licensed, publicly released speech datasets, including synthetic and
human-recorded audio. Internal processing comprised curation, filtering, and
automatic re-annotation rather than new participant data collection.
Evaluation used public benchmarks. No speaker re-identification was attempted.

\bibliographystyle{IEEEbib}
\bibliography{strings,refs}

\end{document}